\documentclass[sigconf,authorversion,nonacm]{acmart}

\hypersetup{final}
\usepackage[algoruled,boxed,lined]{algorithm2e}
\usepackage{grffile} 
\usepackage{xcolor}
\usepackage{supertabular} 
\usepackage{cleveref}
\usepackage{multirow}
\usepackage{listings}
\usepackage{tabularx}
\usepackage{newfloat}
\DeclareFloatingEnvironment[
   fileext=lol,
   name=Listing
]{listing}

\usepackage{subcaption}
\DeclareCaptionSubType{listing}

\usepackage{mathtools} 
\usepackage{enumitem}

\makeatletter
\def\@copyrightspace{\relax}
\makeatother

\begin{document}

\title{Temporal Knowledge Distillation for Time Sensitive Financial Services Applications}

\author{Hongda Shen}
\affiliation{
  \institution{University of Alabama in Huntsville}
  \country{}
}
\email{hs0017@alumni.uah.edu}

\author{Eren Kurshan}
\affiliation{
  \institution{Princeton University}
  \country{}
}
\email{ekurshan@princeton.edu}



\begin{abstract}
Detecting anomalies has become an increasingly critical function in the financial service industry. Anomaly detection is frequently used in key compliance and risk functions such as financial crime detection, fraud, and cybersecurity. The dynamic nature of the underlying data patterns, especially in adversarial environments like fraud detection, poses serious challenges to the machine learning models. Keeping up with the rapid changes by retraining the models with the latest data patterns introduces pressures in balancing the historical and current patterns while managing the training data size. Furthermore, the model retraining times raise problems in time-sensitive and high-volume deployment systems, where the retraining period directly impacts the model's ability to respond to ongoing attacks in a timely manner.  In this study, we propose a temporal knowledge distillation-based label augmentation approach, (TKD), which utilizes the learning from older models to rapidly boost the latest model and effectively reduces the model retraining times to achieve improved agility. Experimental results show that the proposed approach provides advantages in retraining times, while improving the model performance.
\end{abstract}

\settopmatter{printfolios=true} 
\pagestyle{plain}
\maketitle

\section{Introduction}
\label{sec:intro}
Machine learning and artificial intelligence solutions have been widely used in the financial services industry. The applications of AI/ML range from trading to consumer and business credit decisions, financial crime detection, mobile banking, and authentication systems \cite{wef2020}. These use cases also span almost all critical business functions from business operations to risk management \citep{Heaton16, Neurips19}. Overall, AI and machine learning solutions have a significant impact on the performance and operations of today's financial services firms.

Financial firms frequently use machine learning to detect anomalies in cybersecurity applications, fraud detection, compliance, and financial crime detection \cite{AnandakrishnanKS2017}. Among these, there is a sizable list of mission-critical applications, each of which requires effective and timely detection as well as response to anomalous events promptly \cite{AnomalyFinance}. 

The number of models used in high-volume and time-sensitive applications is growing due to the recent trends in mobile payments and digital banking \cite{EU}. In the past few years, mobile payment systems have experienced unprecedented growth \cite{Zelle2}. Zelle transaction volumes increased by 61\% year-to-year as of 2021 \cite{AmericanBanker},\cite{Zelle}. As the high-speed, high-volume nature of digital transactions attracts financial crime organizations,  digital payments fraud and cybercrimes have rapidly become among the top challenges in financial firms.   In digital payments, mobile P2P fraud experienced a 733\% increase from 2016 to 2019. Similarly, account takeover fraud (ATO) grew by 72\% just in one year, from 2018 to 2019 \cite{Forbes} and rose over 200\% in 2021 according to industry reports \cite{techrepublic}. 

One of the grand challenges machine learning models face in these use cases is the dynamic and adversarial nature of the detection process. Unlike naturally stable datasets used in other application areas of AI/ML, fraud and anomaly detection systems experience frequent and rapid pattern changes \cite{MarfaingG2018}, \cite{Mckinsey}. The pattern changes occur in both (i) \textit{normal events}, as in changes in the non-fraud transactions and customer behavior, as well as in (ii)  \textit{anomalous events}, where perpetrators implement new fraud tactics. As an example, in account takeover fraud (which is characterized by perpetrators gaining access to the customers' account and draining the funds across multiple channels  \cite{EU},\cite{ATO_News}), fraud tactics are known to show rapid changes. In some instances, the patterns change from one popular tactic to the next in a matter of hours. This raises serious concerns for industry-standard machine learning solutions that rely on supervised learning. In payment fraud detection systems, which process and score millions of transactions every day with millisecond response time SLAs, the underlying modeling challenges become more pronounced \cite{Mckinsey_cyber},\cite{Deloitte}, \cite{WorldBank}.

In high-volume and dynamic application environments, such as payment fraud detection, models face serious dilemmas: 
\begin{itemize}[leftmargin=0.3cm] 
\item
\textit{Data Size \& Balance}: Retraining the models with the new patterns is a common approach to improve the model performance for recent data patterns. However, it often degrades the performance of historical patterns that repeat. Excluding the historical patterns causes retention challenges. Yet, continuously extending the training data set with additional data causes size and training time issues \cite{TowardsDS}. This highlights the need to balance the historical and recent data while preventing uncontrolled growth in the training data. Similarly, in anomaly detection cases model retraining aims to (i) capture all anomalous patterns and (ii) balance the anomalous/normal event percentages in the training data to achieve the highest model performance. The combination of both goals often causes the machine learning models to use larger data sizes to deal with highly dynamic environments. 
\vspace{2pt}
\item 
\textit{Training Times \& Model Agility}: Similarly, the amount of time required to retrain the models is a serious challenge in the deployment of time-sensitive production models. Models need to be rapidly updated and moved back to production systems to prevent further financial and cybersecurity damage. In response, numerous techniques have been explored to reduce the training times. Hardware acceleration using graphical processing units (GPU) \cite{GPU}, \cite{Amazon}, field programmable gate arrays (FPGA) \cite{Cong}\cite{FPGASurvey}, memory optimizations \cite{IBM}, vectorization and other techniques have been proposed \cite{Rice}, \cite{CVR}, \cite{IBMKailash}. While the hardware and system optimizations provide benefits, optimizing the models themselves for training time is equally important in improving the overall solution.  

\end{itemize}

This paper explores a novel supervised learning approach to tackle these key challenges by providing a way to update models faster,  while balancing the current and historical datasets in high-volume and time-sensitive financial services applications. The proposed technique, Temporal Knowledge Distillation (TKD), transfers knowledge from the historical data to boost the model performance through data label augmentation. It aims to balance the data to enhance the model's robustness. Further, it improves the training time for agile response in adversarial use cases, such as fraud detection and account takeover. The paper explores a fraud detection use case to analyze the effectiveness of the proposed approach. However, the approach can be broadly applied to a wide range of financial services applications due to the prominence of high-volume and time-sensitive applications in the financial services systems.

The paper is organized as follows: \Cref{sec:RelatedWork} discusses the related work in knowledge distillation; \Cref{sec:TKD} outlines the TKD temporal knowledge distillation approach; \Cref{sec:ExpAnalysis} describes the experimental analysis setup; \Cref{sec:ExpResults} overviews the experimental results; finally \Cref{sec:Conclusions} discusses the conclusions.


\section{Related Work}
\label{sec:RelatedWork}
The concept of Knowledge Distillation (KD) was explored by a number of researchers  \cite{BuciluaC2006,BaJ2014,HintonG2015,UrbanG2016,FurlanelloT2018}. Initially, the goal of KD was to produce a compact student model that retains the performance of a more complex teacher model that takes up more space and/or requires more computation to make predictions. \textit{Dark Knowledge} \cite{HintonG2015}, which includes a softmax distribution of the teacher model, was first proposed to guide the student model. 

Recently, the focus of this line of research has shifted from model compression to label augmentation which can be considered a form of regularization using \textit{Dark Knowledge}. In \cite{FurlanelloT2018}, a chain of retraining models, parameterized identically to their teachers, \textit{Born Again Network} (BAN), was proposed. The final ensemble of all trained models can outperform their teacher network on computer vision and NLP tasks. Additionally, \cite{FurlanelloT2018} investigated the importance of each term to quantify the contribution of dark knowledge to the success of KD. 

Following this direction of research, self distillation has emerged as a new technique to improve the classification performance of the teacher model rather than merely mitigating computational or deployment burden. Label refinery \cite{BagherinezhadH2018} iteratively updates the ground truth labels after cropping the entire image dataset and generates a set of informative, collective, and dynamic labels, from which one can learn a more robust model. In another related study, \cite{RomeroA2014} aimed to compress models by approximating the mapping between hidden layers of the teacher and the student models, using linear projection layers to train relatively narrower students. 

KD has already gained success in many applications in a wide range of domains. In \cite{YangZZ2019}, authors have proposed to distill the knowledge of essence in an ensemble of models to a single model that needs much less computation to deploy for building a speech recognition system. Similarly, KD was utilized in computer vision domain: object detection \cite{ChenCY2017}, deepfake detection \cite{KimTW2021}, image super-resolution \cite{ZhangCC2021}. \cite{SanhDC2019} proposed a method to pre-train a smaller general-purpose language representation model, called DistilBERT based on Knowledge Distillation, which can reduce the size of a large BERT NLP model by 40\%. In the emerging field of federated learning, KD has started to be used to tackle some challenges such as user heterogeneity issue \cite{ZhuHZ2021} and expensive model payload during communication \cite{SeoPO2020}. Besides the traditional response-based distillation, feature-based distillation \cite{ChenMZ2020} has recently gained attention from researchers with an attempt to leverage knowledge from the intermediate layers in addition to the output layer \cite{GouYM2020}. As KD shifts its focus from model compression to domain adaption, it has shown its  potential in handling heterogeneous data sources. Our work is motivated by this trend and shares some conceptual similarities with \cite{FarhadiY2019} for both attempt to distill knowledge temporally. However,  \cite{FarhadiY2019} utilizes temporal correlations across video frames while the proposed work in this paper mainly tackles the dynamic nature of anomaly detection use case.

The proposed TKD label augmentation incorporates \textit{Dark Knowledge} from previously trained models, which have been trained with different time ranges to augment the labels of the latest dataset. This new knowledge enables the transfer of learning from historical patterns extracted by experienced \textit{experts}. With the assistance of their expertise, the new model sees performance improvement without having the historical datasets in its training. This enables more effective detection of anomalous events, and streamlines model retraining and deployment in a time-sensitive scenario.

\section{Temporal Knowledge Distillation}
\label{sec:TKD}
Consider the classical classification setting with a sequence of training datasets corresponding to $N$ different time frames consisting feature vectors: $X_t$ and labels $Y_t$ where $t=0,1,...N$. For traditional supervised learning algorithms, a model is trained on $\{X_{<t},Y_{<t}\}$ for each time frame. Naturally, the size of $\{X_{<t},Y_{<t}\}$ increases as time passes. TKD leverages the outputs generated by previously trained models $M_{<t}$  prior to each time frame $t$ instead of including historical data in the training directly. These outputs are used to augment labels of the latest dataset and construct a regularizer to the conventional loss function. For time frame $t$, the training dataset will be $\{X_{t},Y_{t}\}$ only and the general form loss function to optimize in the training becomes:

\begin{equation}
\label{eq:loss}
Loss_{t} =  \alpha\cdot CE(Y_t, y_t)+ (1-\alpha)\cdot\underset{K \leq i \leq t-1}{AGG} \left[ KL(O_{i,t},y_t))\right]
\end{equation}
where $O_{i,t}$ and $y_t$ represents model $M_i$ output on data $X_t$ and model output at the current time frame, respectively.  $CE$ and $KL$ are Cross-Entropy and Kullback–Leibler divergence. With this second term in the loss function Eq. \ref{eq:loss}, existing ground truth labels are augmented by the \textit{experienced experts}. Typically,  KL divergence is used to minimize the discrepancy between the logits outputs from the
teacher model and the student model, respectively.

$AGG[\cdot]$ serves as an aggregating function which collects `vote` from each \textit{experienced expert}. There are many options for this aggregating function e.g. $max()$, $sum()$, $mean()$ and etc. and in this work $mean()$ is selected after some empirical experiments. The coefficient $\alpha$ is used for balancing the label augmentation term and the cross-entropy on new data. As $\alpha$ gets closer to one, the training for the target model depends more on the ground-truth labels of the new data and this approach gradually converges to a classic classification without any regularizer. For simplicity, $\alpha$ is set to 0.5 in this paper.

As the number of models increases over time, the historical models, whose underlying training data patterns have changed provide increasingly less meaningful information on the recent anomaly patterns. Thus, including them in the training may not provide further performance gain for retraining and possibly deteriorate the performance. To reduce the negative impact of this distribution shift, we use parameter $K$ to determine which model to start with and truncate all the previous models prior to the current one. In this study we used an empirical approach to determine $K$. With all the specific setups, the loss function used can be simplified as follows:

\begin{equation}
\label{eq:simple_loss}
Loss_{t} =  CE(Y_t, y_t)+ \frac{1}{t-K}\sum_{i=K}^{t-1}  KL(O_{i,t},y_t))
\end{equation}


\begin{figure}[h]
  \centering
  \includegraphics[scale=0.65]{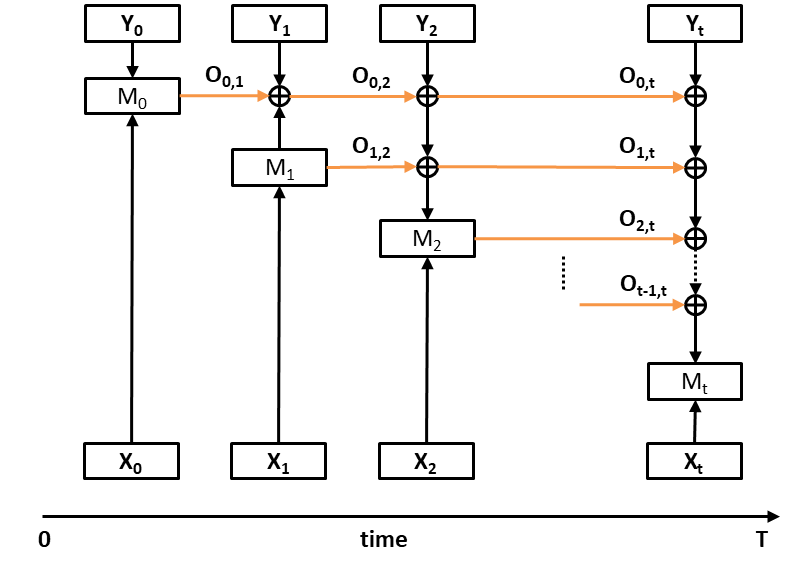}
  \caption{Architecture of Label Augmentation via Time-based Knowledge Distillation (TKD).}
\label{fig:diagram}
\vspace{0em}
\end{figure}


Fig. \ref{fig:diagram} illustrates the architecture of TKD. For the first time frame $t=0$, a model $M_0$ is trained on dataset $\{X_{0},Y_{0}\}$. Then, for each of the following time frames (depending on the specific retraining schedule), a new identical model $M_t$ is trained from, $O_{i,t}$ the supervision of previous models $M_{<t}$ by using Eq. \ref{eq:simple_loss}. Auxiliary soft labels (outputs) from the previous models are highlighted in orange in Fig. \ref{fig:diagram}. Note that this figure depicts the $K=0$ case. When the number of historical models becomes large, certain truncation can be applied based on prior knowledge of each historical model.


\section{Experimental Analysis}
\label{sec:ExpAnalysis}

\subsection{Dataset}
In order to evaluate the effectiveness of TKD in time-sensitive anomaly detection applications, we use a fraud detection use case. This section provides the experimental simulation analysis setup for TKD using an open-source anomaly detection dataset \cite{IEEE_CIS_Dataset} based on telecommunications industry card-not-present digital payment transactions. As in almost all the anomaly detection problems, the negative class in this data set takes a very small portion of the total transactions. For the experimental analysis, we extracted 6 months of data with the labels included. The first day of this data set is assumed to be November 1st, 2017 \cite{IEEE_CIS_Dataset_Timeframe_Analysis}. The start date was used to facilitate data segmentation and does not impact the model performance. Details about the positive/negative samples distribution of the experimental dataset can be found in Table \ref{tab:dataset}.

\begin{table}[h!]
\caption{Overview of the transaction and fraud distribution in the experimental dataset.}
\label{tab:dataset}
\scalebox{1.0}{
\begin{tabular}{lrr}
\toprule
\multicolumn{1}{c}{Month} & \multicolumn{1}{c}{Nonfraud \#} & \multicolumn{1}{c}{Fraud \#} \\ \hline
Nov-17 & 130,937 & 3,401 \\ 
Dec-17 & 88,821 & 3,689 \\ 
Jan-18 & 95,398 & 3,939 \\ 
Feb-18 & 84,785 & 3,571 \\ 
Mar-18 & 83,723 & 2,949 \\ 
Apr-18 & 83,577 & 2,995 \\ 
\hline
Total & 567,241	& 20,544\\
\bottomrule
\end{tabular}
}
\end{table}

\subsection{Experimental Setup}
\label{section:exp_setup}
Model training period begins with November 2017 and gradually incorporates additional months to the training period to approximate an adversarial fraud detection environment. Also, a one-month delay policy is assumed for data labeling. This accounts for the fraud claim submission process, which is a hybrid of digital and manual, as well as labeling the reported transactions. Accordingly, the testing period started from January to April 2018 and with one reduced month from the previous run. 
Table \ref{tab:exp_runs} shows further details of the experiment periods. In total, there are four experiment periods used for the experimental analysis. For each period, TKD trains the model with the most recent data only (they are highlighted in bold and red in Table \ref{tab:exp_runs}). Note that Period 0 does not have any pre-trained model to run TKD.

\begin{table}[h]
  \caption{Experimental period details.}
  \label{tab:exp_runs}
  \centering
  \scalebox{0.8}{
  \begin{tabular}{clcr}
    \toprule
    Period \#  & Training Period   & No-label Period & Testing Period \\
    \midrule
    0 & Nov. & Dec. & Jan. + Feb. + Mar. + Apr. \\
    1 & Nov. + \textbf{\textcolor{red}{Dec.}} & Jan. & Feb. + Mar. + Apr.\\
    2 & Nov. + Dec. + \textbf{\textcolor{red}{Jan.}} & Feb. & Mar. + Apr. \\
    3 & Nov. + Dec. + Jan. + \textbf{\textcolor{red}{Feb.}} & Mar. & Apr. \\
    \bottomrule
  \end{tabular}}
\end{table}

Prior to the model training, categorical features are encoded using \textit{one-hot encoding}. \textit{log10} transformation is used on continuous variables to limit their value ranges. Further details on data preprocessing can be found in Table \ref{tab:dataset_preprocessing} in the Appendix. For performance comparisons, Area Under Precision-Recall Curve (AUPRC or AUC-PR) was selected as the primary metric to compare the classification performance results. AUPRC has been shown to be a stronger metric for performance and class separation than the traditional Area Under Receiver Operating Curve (AUROC or AUC-ROC) in highly imbalanced binary classification problems such as anomaly detection use cases \cite{JesseD2006,SaitoT2015}.

\subsection{Algorithm Comparison}
\label{section:candidates}
In order to cover the model types for fraud detection use cases both neural networks and decision tree-based algorithms were used in the experimental analysis \cite{ShenK2020}. Similarly, ensemble techniques \cite{Zhou2012} have been widely reported in many classification and fraud detection applications \cite{ForoughM2021}. Therefore, an ensemble of neural network and tree-based model was incorporated in the experimental analysis. Altogether, the analysis aims to compare the performance of the baseline neural networks, tree-based models and ensemble models along with their corresponding TKD versions:
\begin{itemize}
    \item [(i)] \textit{MLP}: A Multi-layer Perceptron has been trained on labeled data to serve as the baseline. Implementation details of the MLP has been provided in Table \ref{tab:mlp} in Appendix.
    \item [(ii)] \textit{XG}: Xgboost is commonly used for its high efficiency and performance \cite{ChenT2016}.  It has been widely used to model tabular data (from Kaggle competitions to industrial applications). The specific set of hyper-parameters for this study were determined using grid search and provided in Table \ref{tab:xg} in Appendix. 
    \item [(iii)] \textit{MLP-XG}: An ensemble of baseline Xgboost and MLP via averaging outputs of both models 
    \item [(iv)] \textit{MLP-TKD}: Label Augmented version of the Multi-Layer Perceptron Model through Temporal Knowledge Distillation (TKD) using historical ensemble models. 
    \item [(v)] \textit{XG-TKD}: Label Augmented version of GXBoost through TKD (Temporal Knowledge Distillation) using historical ensemble models. 
    \item [(vi)]  \textit{MLP-XG-TKD}: Label Augmented version of MLP-XG-ensemble approach through TKD (Temporal Knowledge Distillation) using historical ensemble models.
\end{itemize}

It is worth highlighting that although most of the related work focused on distilling knowledge from a neural network to another, having both MLP and XG allows the study to explore whether it is also viable to transfer knowledge between two heterogeneous architectures (e.g. a neural network and a tree). Moreover, with the historical models (teacher models) being ensembles of heterogeneous architectures, the study tries to explore if the student network can still benefit from such a knowledge source. 

A supervised binary classification approach was used to train the fraud detection classifier, where each algorithm was run 10 times for each training time period. AUPRC value for each run is collected and the average of all the 10 collected values is recorded as the final performance measure.

\section{Experimental Results}
\label{sec:ExpResults}
This section presents the performance comparisons between baseline machine learning models and their TKD counterparts. Both one month and consecutive months results as described in the Section are presented \ref{section:exp_setup}). Next, we analyze the model training time over the entire 6 months experiment period to show the training time characteristics of TKD compared to the baseline.

\subsection{Model Performance}
As shown in Table \ref{tab:exp_runs}, experiment period 3, which has the highest number of months in its training period and the most historical models for TKD, was used as the primary test period. Note that the focus is to evaluate TKDs effectiveness rather than finding the best performing model. Therefore, the three baseline models (MLP, XG, MLP-XG) were paired  with their TKD versions. AUPRC comparisons were made between regular training and TKD training. 

\begin{figure}[h]
  \centering
  \includegraphics[scale=0.465]{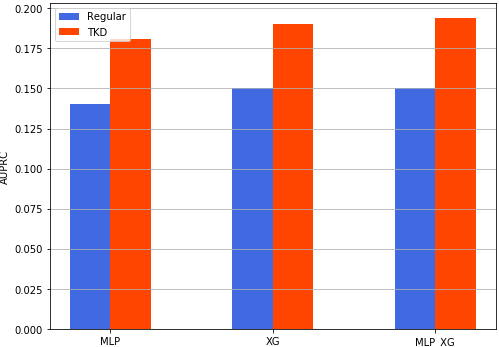}
    \vspace{-1em}
  \caption{AUPRC for MLP, XG, MLP-XG pairs over Period 3.
}
\vspace{0em}
\label{fig:aucpr_period_3}
\end{figure}

\begin{table}[h!]
  \caption{Relative AUPRC difference of TKD methods against their base models.}
  \label{tab:}
  \centering
  \scalebox{1}{
  \begin{tabular}{crr}
    \toprule
    Model \# &  $\Delta$ & Relative Improvement (\%)\\
    \midrule
    MLP & 0.0402 & 28.59\% \\
    XG & 0.3097 & 26.39\% \\
    MLP\_XG & 0.0435 & 28.98\% \\
    \bottomrule
  \end{tabular}}
\label{tab:relative_auprc}
\end{table}

Fig. \ref{fig:aucpr_period_3} shows the AUPRC for each model candidate listed in Section \ref{section:candidates} over period 3. For each pair,  TKD versions produce significantly higher AUPRC hence higher performance in fraud detection. Metric $\Delta$, defined as the absolute gain of AUPRC of TKD version over the baseline, is presented in Table \ref{tab:relative_auprc}. In addition, the relative percentage of the metric is also shown to better quantify performance improvement by TKD. Similar to what is observed in Fig. \ref{fig:aucpr_period_3}, TKD method consistently improves the model performance over the baseline regardless of the model type.

\begin{figure}[h]
  \centering
  \includegraphics[scale=0.43]{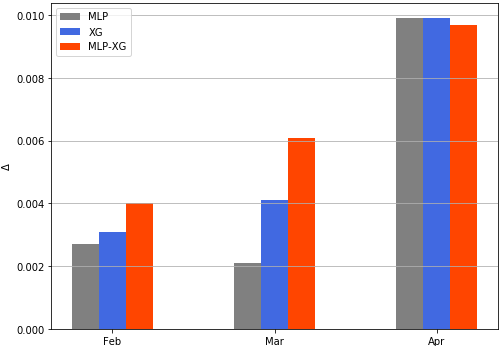}
  \caption{AUPRC $\Delta$ for MLP, XG, and MLP-XG pairs over Period 1.}
\vspace{0em}
\label{fig:aucpr_period_1}
\end{figure}

To assess TKD's impact on model stability over time, Period 1 was used as the primary time period (as there are three months for testing in Period 1 and one historical model available from Period 0 to enable TKD). It is important to note that AUPRC cannot be compared across datasets since they might have different positive sample ratios. Hence, in this experiment, $\Delta$ was used as the performance metric for each testing month from February to April 2018 in Fig. \ref{fig:aucpr_period_1}.

\begin{figure}[h]
  \centering
  \includegraphics[scale=0.43]{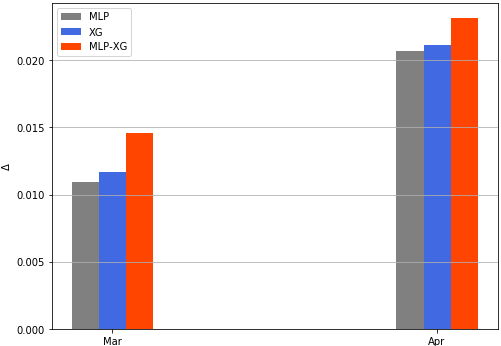}
  \caption{AUPRC $\Delta$ for MLP, XG, and MLP-XG pairs over Period 2.
}
\vspace{0em}
\label{fig:aucpr_period_2}
\end{figure}

Overall, Fig. \ref{fig:aucpr_period_1} shows that  TKD produces the highest performance gain in April 2018, which is the last month of the testing period. This indicates that TKD provides durable robustness, which is of interest in stabilizing fraud detection models performance over time in a dynamic production environment. Interestingly, even though $\Delta$ values are quite close for the three model pairs, it appears that the ensemble of different model types benefits the most compared to the other models. Similar observations can be made for Period 2; although it only has two months for testing. Please see Fig. \ref{fig:aucpr_period_2} for the results of Period 2.

\subsection{Model Training Time}
As discussed in \Cref{sec:intro}, the reactive model retraining period in dynamic environments introduces significant challenges in terms of data size and  the time to retrain the model.  The pressure to push the retrained model back into production to prevent ongoing attacks and crime patterns translates to some critical goals: (i) reducing retraining times (ii) using well-balanced data with historical and current data patterns without overgrowing the training data size (which in turn affects goal (i) as well). 

Due to the imbalance in the classes, fraud detection models use over-sampling for fraud cases and under-sampling for non-fraud transactions to achieve a target fraud/non-fraud composition. In order to achieve the highest-performance model all fraud cases are used in model training. This, in turn, dictates the number of non-fraud cases to be used in the model training. As a result, with emerging fraud patterns training data sizes practically grow in reactive retraining cases. 

Most modern machine learning algorithms (including neural networks and tree-based models) are trained in batch optimization fashion, for which additional model training data translates to improved performance \cite{DevriesT2017,SahinS2018}. Therefore, the training time is typically longer in higher performing models that use larger datasets.

\begin{figure}[h]
  \centering
  \includegraphics[scale=0.49]{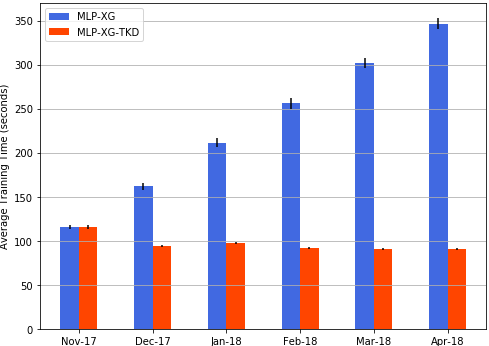}
  \caption{Average model training time comparison between MLP-XG and MLP-XG-TKD.
}
\vspace{0em}
\label{fig:runtime}
\end{figure}

Based on the performance analysis, MLP-XG ensemble and MLP-XG-TKD were identified as the highest performance models. Fig.\ref{fig:runtime} shows the average model training time in seconds for MLP-XG and MLP-XG-TKD over 10 repeated runs from November 2017 to April 2018. A machine with Intel (R) Core (TM) i7-6700HQ CPU at 2.6GHz, 16GB RAM and NVIDIA GTX 960M GPU was used for this comparison. 

MLP-XG was trained with cumulative time periods of data (similar to Table \ref{tab:exp_runs}), while TKD training only included the month itself without any historical data. An extended version of the time range up to Apr-18 was used to better illustrate the training time benefits over time. Since TKD provides a way to transfer knowledge hidden in the historical models, without explicitly training on historical data, the average training time only depends on the size of the latest dataset. On the other hand, traditional supervised learning techniques including both MLP and XG require all historical data in their training, which leads to super-linear increases in the training time. Fig. \ref{fig:runtime} shows the average training times for MLP-XG (blue) and MLP-XG-TKD (red). Both methods take the same time to run at the beginning because no historical models are available for TKD. 

Gradually, with more data used in MLP-XG training, its training time increases, while that of MLP-XG-TKD remains approximately the same. MLP-XG-TKD provides faster training consistently over Dec-17 through Apr-18.  Over the 6 months of experimentation period, the average training time was reduced by 58.5\% with up to 3.8x improvement in Apr-18. 

It is important to note that the training time advantage of TKD shown in this experiment translates to significant impact in real-life implementations with large production data sets, yielding reduced training time cost, and resources. This, in turn, yields improved computational cost and agility of responses in adversarial production environments in time-sensitive tasks. TKD provides the opportunity to boost the performance of the baseline models as well as the ensemble models.

\section{Conclusions}
\label{sec:Conclusions}
This study proposes a label augmentation algorithm, Temporal Knowledge Distillation (TKD), for time-sensitive financial anomaly detection applications. This technique aims to provide a new way to boost the model performance by incorporating a wider range of patterns including older and newer patterns without causing unmanageable increases in the data size. 
Furthermore, it  minimizes the model retraining times compared to the baseline models. In adversarial and time-critical use cases, such as cybersecurity and payment fraud detection applications, this yields significantly higher agility and more effective response capabilities to attacks.  Despite the recent advancements in acceleration techniques, model retraining times remain as a challenge in deploying AI/ML models in production systems. TKD delivers an alternative approach to model retraining in time-sensitive and high-volume applications, which provides key benefits to the overall success of the deployment systems.

\bibliography{bib}
\bibliographystyle{ACM-Reference-Format}

\section*{Appendix}
\begin{table}[hb]
  \caption{Dataset Pre-processing Details}
  \label{tab:dataset_preprocessing}
  \centering
  \scalebox{0.65}{
  \begin{tabular}{lllll}
    \toprule
    Raw feature  & Type   & Encoding & Null value & Notes \\
    \midrule
    TransactionAmt & Continuous & $log10()$  & - &   -   \\
    dist1 & Continuous & $log10()$ & $-0.001$ &  -    \\
    dist2 & Continuous & $log10()$  & $-0.001$ &  - \\
    ProductCD & Categorical & One hot  & - & - \\
    card4 & Categorical & One hot  & NA  & -\\
    card6 & Categorical & One hot  & NA  & -\\
    M1-M9 & Categorical & One hot  & NA  & -\\
    device\_name & Categorical & One hot  & NA  & ``Others" if frequency $<$ 200\\
    OS & Categorical & One hot  & NA  & -\\
    Browser & Categorical & One hot  & NA  & ``Others" if frequency $<$ 200\\
    DeviceType & Categorical & One hot  & NA  & -\\
    \bottomrule
  \end{tabular}}
\end{table}

\begin{table}[hb]
  \caption{Multi-layer Perceptron Architecture}
  \label{tab:mlp}
  \centering
  \scalebox{0.77}{
  \begin{tabular}{llll}
    \toprule
    Layer  & \# Neurons  & Activation function & Parameter \\
    \midrule
    Dense & 400 & RELU & -   \\
    BatchNormalization & - & - & -   \\
    Dropout & - & -  & keep\_prob = 0.5 \\
    Dense & 400 & RELU  & -  \\
    Dropout & - & -  & keep\_prob = 0.5\\
    Dense (Output) & 2 & Softmax & -\\
    \midrule
    learning rate & - & - & 0.01\\
    Batch size & - & - & 512\\
    \bottomrule
  \end{tabular}}
\end{table}

\begin{table}[hb]
  \caption{Xgboost Hyperparameters}
  \label{tab:xg}
  \centering
  \scalebox{0.8}{
  \begin{tabular}{ll}
    \toprule
    Name  & Value  \\
    \midrule
    colsample\_bytree & 0.8   \\
    gamma & 0.9   \\
    max\_depth & 3 \\
    min\_child\_weight & 2.89  \\
    reg\_alpha & 3\\
    reg\_lambda & 40\\    
    subsample & 0.94\\
    \midrule
    learning\_rate & 0.1\\
    n\_estimators & 200\\
    \bottomrule
  \end{tabular}}
\end{table}

\end{document}